\documentclass[10pt,twocolumn,letterpaper]{article}

\usepackage{cvpr}
\usepackage{times}
\usepackage{epsfig}
\usepackage{graphicx}
\usepackage{amsmath}
\usepackage{amssymb}
\usepackage{amsthm,mathrsfs,amsfonts,dsfont}
\usepackage{multirow}
\usepackage{longtable}
\usepackage{supertabular}
\usepackage{rotating}
\usepackage{subfigure}

\usepackage[breaklinks=true,bookmarks=false]{hyperref}

\cvprfinalcopy 

\def\cvprPaperID{2516} 

\begin{document}

\title{More is Less: A More Complicated Network with Less Inference Complexity\thanks{This paper was accepted by the IEEE CVPR 2017}}

\author{Xuanyi Dong$^{1}$\thanks{This work was done when Xuanxi Dong was an Intern at 360 AI Institute.}, Junshi Huang$^{2}$, Yi Yang$^{1}$, Shuicheng Yan$^{2,3}$\\
$^{1}$CAI, University of Technology Sydney, $^{2}$360 AI Institute, $^{3}$National University of Singapore\\
{\tt\small dongxuanyi888@icloud.com; huangjunshi@360.cn}\\ 
{\tt\small yi.yang@uts.edu.au; yanshuicheng@360.cn}
}

\maketitle

\begin{abstract}

In this paper,
we present a novel and general network structure towards accelerating the inference process of convolutional neural networks,
which is more complicated in network structure yet with less inference complexity.
The core idea is to equip each original convolutional layer with another low-cost collaborative layer (LCCL),
and the element-wise multiplication of the ReLU outputs of these two parallel layers produces the layer-wise output.
The combined layer is potentially more discriminative than the original convolutional layer,
and its inference is faster for two reasons:
1) the zero cells of the LCCL feature maps will remain zero after element-wise multiplication,
and thus it is safe to skip the calculation of the corresponding high-cost convolution in the original convolutional layer;
2) LCCL is very fast if it is implemented as a $1 \times 1$ convolution or only a single filter shared by all channels.
Extensive experiments on the CIFAR-10, CIFAR-100 and ILSCRC-2012 benchmarks show that our proposed network structure
can accelerate the inference process by 32\% on average with negligible performance drop.

\end{abstract}

\section{Introduction}

\begin{figure}[!t]
\begin{center}
\includegraphics[width=0.48\textwidth]{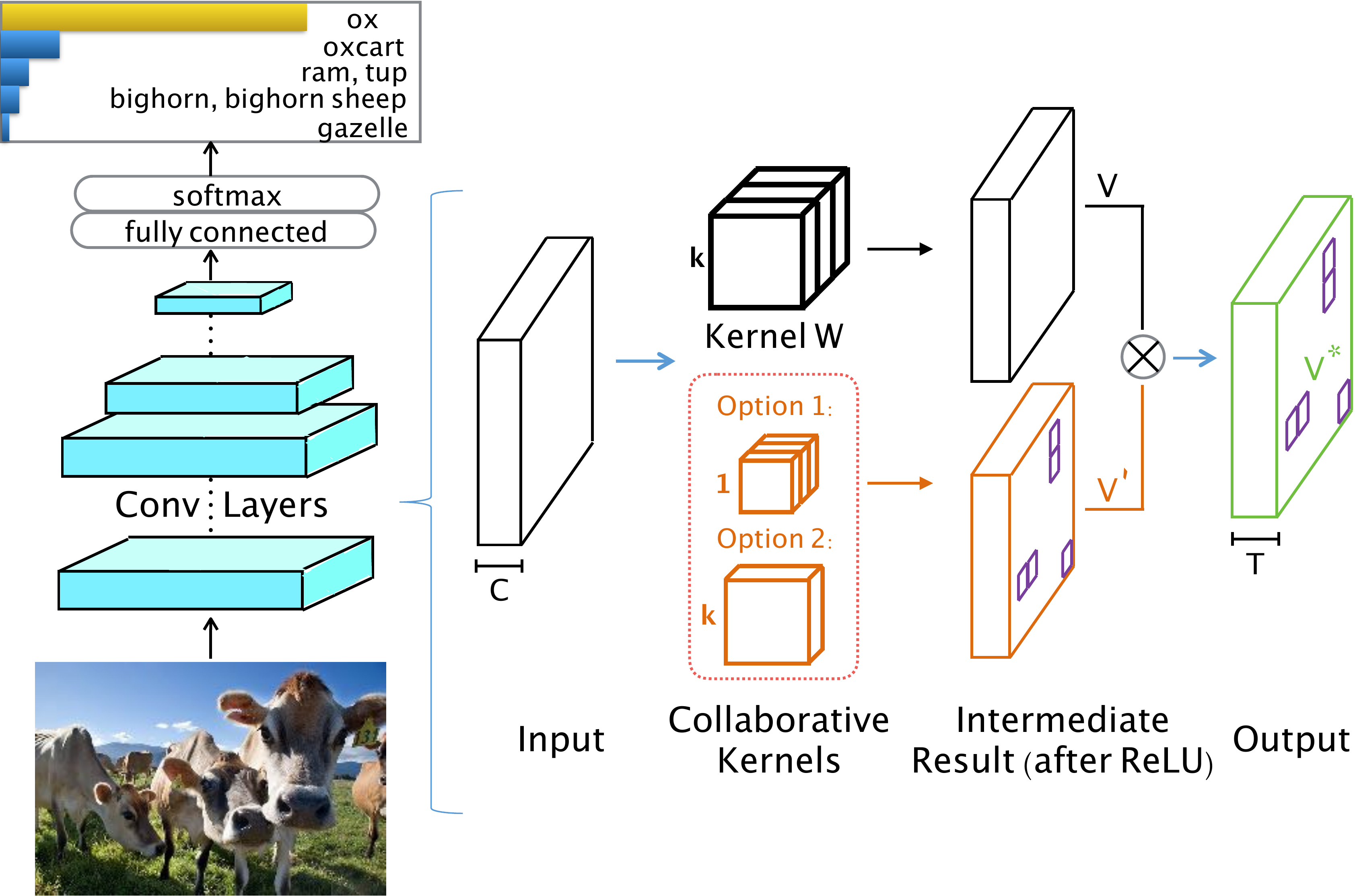}
\end{center}
\caption{Basic acceleration block. 
The orange panel in the figure shows two different kinds of low-cost collaborative kernels.
One uses $1 \times 1$ convolution, and
the other uses shared kernels~($W_i^{'} = W_j^{'}$ for $i,j \in [1, T]$).
The black response map represents the output of the original convolutional layer with the kernel $W$,
and the orange response map is generated by the low-cost collaborative layer.
The purple cells represent the zero elements, of which the calculation of corresponding positions can be skipped in the original convolutional layer.
We apply element-wise multiplication on the activated response maps from the original convolutional layer and low-cost layer to generate the final results of this basic acceleration block.}
\label{fig:basic_framework}
\end{figure}

Despite the continuously improved performance of convolutional neural networks (CNNs)~
\cite{chatfield2014return,han2015deep,krizhevsky2012imagenet,lin2013network,simonyan2014very,szegedy2015going}, their
computation costs are still tremendous.
Without the support of high-efficiency servers,
it is hard to establish CNN models on real-world applications.
For example, to process a $224 \times 224$ image,
AlexNet~\cite{krizhevsky2012imagenet} requires 725M FLOPs with 61M parameters,
VGG-S~\cite{chatfield2014return} involves 2640M FLOPs with 103M parameters,
and GoogleNet~\cite{szegedy2015going} needs 1566M FLOPs with 6.9M parameters.
Therefore, to leverage the success of deep neural networks on mobile devices with limited computational capacity,
accelerating network inference has become imperative.

In this paper, we investigate the acceleration of CNN models based on the observation that the response maps of many convolutional layers are usually sparse after ReLU~\cite{montufar2014number} activation.
Therefore, instead of fully calculating the layer response,
we can skip calculating the zero cells in the ReLU output and only compute the values of non-zero cells in each response map. 
Theoretically, the locations of zero cells can be predicted by a lower cost layer.
The values of non-zero cells from this lower-cost layer can be collaboratively updated by the responses of the original filters.
Eventually, the low-cost collaborative layer (LCCL) accompanied by the original layer constitute the basic element of our proposed low-cost collaborative network (LCCN).

To equip each original convolutional layer with a LCCL,
we apply an element-wise multiplication on the response maps from the LCCL and the original convolutional layer,
as illustrated in Fig.~\ref{fig:basic_framework}.
In the training phase, this architecture can be naturally trained by the existing stochastic gradient descent (SGD) algorithm with backpropagation.
First we calculate the response map $V^{'}$ of the LCCL after the activation layer,
and use $V^{'}$ to guide the calculation of the final response maps.

Despite the considerable amount of research where a sparse-based framework is used to accelerate the network inference,
\eg~\cite{figurnov2015perforatedcnns,graham2014spatially,lebedev2015fast,li2016pruning,liu2015sparse},
we claim that LCCN is unique.
Generally, most of these sparsity-based methods~\cite{lebedev2015fast,liu2015sparse,soulie2015compression} integrate the sparsity property
as a regularizer into the learning of parameters, which usually harms the performance of network.
Moreover, to further accelerate performance, some methods even arbitrarily zeroize the values of the response maps according to a pre-defined threshold.
Compared with these methods, our LCCN automatically sets the negatives as zero,
and precisely calculates the positive values in the response map with the help of the LCCL.
This two-stream strategy reaches a remarkable acceleration rate
while maintaining a comparable performance level to the original network.

The main contributions are summarized as follows:
\begin{itemize}
  \item We propose a general architecture to accelerate CNNs, which leverages low-cost collaborative layers to accelerate each convolutional layer.
  \item To the best of our knowledge, this is the first work to leverage a low-cost layer to accelerate the network.
  Equipping each convolutional layer with a collaborative layer is quite different from the existing acceleration algorithms. 
  \item Experimental studies show significant improvements by the LCCN on many deep neural networks when compared with existing methods (\eg, a 34\% speedup on ResNet-110).
\end{itemize}

\section{Related Work}


{\bf Low Rank}.
Tensor decomposition with low-rank approximation-based methods are commonly used to accelerate deep convolutional networks. For example,
in~\cite{denton2014exploiting,jaderberg2014speeding}, the authors exploited the redundancy between convolutional filters and used low-rank approximation to compress convolutional weight tensors and fully connected weight matrices.
Yang \etal \cite{yang2015deep} use an adaptive fastfood transform was used to replace a fully connected layer with a series of simple matrix multiplications, rather than the original dense and large ones.
Liu \etal \cite{liu2015sparse} propose a sparse decomposition to reduce the redundancy in convolutional parameters.
In~\cite{zhang2015accelerating,zhang2015efficient}, the authors used generalized singular vector decomposition~(GSVD) to decompose an original layer to two approximated layers with reduced computation complexity.

{\bf Fixed Point}.
Some popular approaches to accelerate test phase computation are based on ``fixed point''.
In~\cite{courbariaux2014training}, the authors trained deep neural networks with a dynamic fixed point format, which
achieves success on a set of state-of-the-art neural networks.
Gupta \etal \cite{gupta2015deep} use stochastic rounding to train deep networks with 16-bit wide fixed-point number representation.
In~\cite{courbariaux2016binarynet,courbariaux2015binaryconnect}, a standard network with binary weights represented by 1-bit was trained to speed up networks.
Then, Rastegari \etal \cite{rastegari2016xnor} further explored binary networks and expanded it to binarize the data tensor of each layer,
increasing the speed by 57 times.

{\bf Product Quantization}.
Some other researchers focus on product quantization to compress and accelerate CNN models.
The authors of \cite{wu2015quantized} proposed a framework to accelerate the test phase computation process with the network parameters quantized and learn better quantization with error correction.
Han \etal \cite{han2015deep} proposed to use a pruning stage to reduce the connections between neurons,
and then fine tuned networks with weight sharing to quantify the number of bits of the convolutional parameters from 32 to 5.
In another work~\cite{hubara2016quantized}, the authors trained neural networks with extremely low precision,
and extended success to quantized recurrent neural networks.
Zhou \etal \cite{zhou2016dorefa} generalized the method of binary neural networks to allow networks with arbitrary bit-width in weights, activations, and gradients.

{\bf Sparsity}.
Some algorithms exploit the sparsity property of convolutional kernels or response maps in CNN architecture.
In~\cite{zhou2016less}, many neurons were decimated by incorporating sparse constraints into the objective function.
In~\cite{graham2014spatially}, a CNN model was proposed to process spatially-sparse inputs,
which can be exploited to increase the speed of the evaluation process.
In~\cite{lebedev2015fast}, the authors used the group-sparsity regularizer to prune the convolutional kernel tensor in a group-wise fashion.
In~\cite{figurnov2015perforatedcnns}, they increased the speed of convolutional layers by skipping their evaluation at some fixed spatial positions.
In~\cite{li2016pruning}, the authors presented a compression technique to prune the filters with minor effects on the output accuracy.

{\bf Architecture}.
Some researchers improve the efficiency of networks by carefully designing the structure of neural networks.
In~\cite{hinton2015distilling}, a simple model was trained by distilling the knowledge from multiple cumbersome models,
which helps to reduce the computation cost while preserving the accuracy. 
Romero \etal \cite{romero2014fitnets} extended the knowledge distillation approach to train a student network,
which is deeper but thinner than the teacher network, by extracting the knowledge of teacher network.
In this way, the student network uses less parameters and running time to gain considerable speedup compared with the teacher network.
Iandola \etal \cite{iandola2016squeezenet} proposed a small DNN architecture to achieve similar performance as AlexNet by only using 50x fewer parameters and much less computation time via the same strategy.

\section{Low-Cost Collaborative Network}

In this section, we present our proposed architecture for the acceleration of deep convolutional neural networks.
First, we introduce the basic notations used in the following sections.
Then, we demonstrate the detailed formulation of the acceleration block and extend our framework to general convolutional neural networks.
Finally, we discuss the computation complexity of our acceleration architecture.

\subsection{Preliminary}

\begin{figure*}[htp]
\begin{center}
\includegraphics[width=0.7\textwidth]{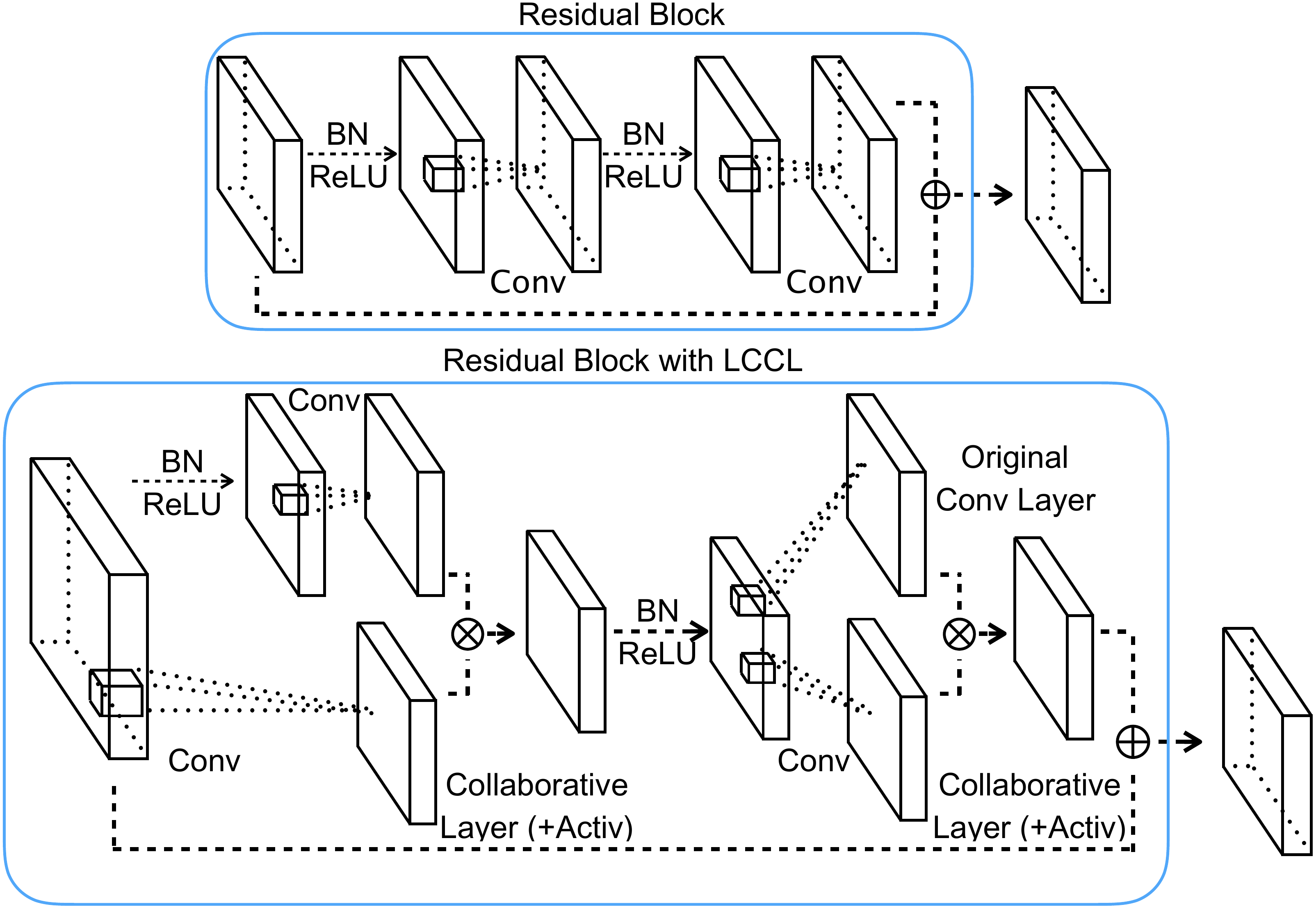}
\end{center}
\caption{Connection strategy of collaborating LCCL with the original convolutional layer.
The top figure shows the pre-activation residual block~\cite{he2016identity};
the bottom figure shows a ``Bef-Aft" connection strategy to speed up the residual block.
``Activ" represents that the collaborative layer is followed by BN and ReLU activation.
The first LCCL receives the input tensor before being activated by BN and ReLU,
and the second one receives the input tensor after BN and ReLU.
(Best viewed in the original pdf file.)} 
\label{fig:position_connect}
\end{figure*}

Let's recall the convolutional operator.
For simplicity, we discuss the problem without the bias term.
Given one convolution layer,
we assume the shapes of input tensor~$U$ and output tensor~$V$ are $X \times Y \times C$ and $X \times Y \times T$,
where $X$ and $Y$ are the width and height of the response map, respectively.
$C$ and $T$ represent the channel number of response map $U$ and $V$.
A tensor~$W$ with size $k \times k \times C \times T$ is used as the weight filter of this convolutional layer.
$V_t(x,y)$ represents the element of $V(x,y,t)$.
Then, the convolutional operator can be written as:

\vspace{-2mm}
{\small
\begin{align}
V_t(x,y) = \sum_{i,j=1}^{k}\sum_{c=1}^{C}W_t(i,j,c)U(x+i-1,y+i-1,c)
\end{align}
}
\vspace{-1mm}

\noindent where $W_t(x,y)$ is the element of $W(x,y,t)$.

In the LCCN, the output map of each LCCL should have the same size as the corresponding convolutional layer,
which means that the shape of tensor~$V^{'}$ is $X \times Y \times T$.
Similarly, we assume the weight kernel of $V^{'}$ is $W^{'}$.
Therefore, the formula of the LCCN can be written as:
{\small
\begin{align}
V^{'}_t(x,y) = \sum_{i,j=1}^{k^{'}}\sum_{c=1}^{C}W^{'}_{t}(i,j,c)U(x+i-1,y+i-1,c)
\end{align}
}
\vspace{-2mm}

\subsection{Overall Structure}

Our acceleration block is illustrated in Fig.~\ref{fig:basic_framework}.
The green block $V^{*}$ represents the final response map collaboratively calculated by the original convolutional layer and the LCCL.
Generally, it can be formulated as:
\vspace{-1mm}
{\small
\begin{align}
V^{*}_t(x,y) =
  \begin{cases} 
   0                             & \text{if } V^{'}_t(x,y) = 0     \\
   V^{'}_t(x,y) \times V_t(x,y)  & \text{if } V^{'}_t(x,y) \neq 0
  \end{cases}
\end{align}
}
\vspace{-1mm}
\noindent where $V$ is the output response map from the original convolutional layer and $V^{'}$ is from LCCL.

In this formula, the element-wise product is applied to $V$ and $V^{'}$ to calculate the final response map.
Due to the small size of LCCL, the computation cost of $V^{'}$ can be ignored.
Meanwhile, since the zero cells in $V^{'}$ will stay zero after the element-wise multiplication,
the computation cost of $V$ is further reduced by skipping the calculation of zero cells according to the positions of zero cells in $V^{'}$.
Obviously, this strategy leads to increasing speed in a single convolutional layer.
To further accelerate the whole network, we can equip most convolutional layers with LCCLs.

\subsection{Kernel Selection}

As illustrated in the orange box in Fig.~\ref{fig:basic_framework}, 
the first form exploits a $1 \times 1 \times C \times T$ kernel ($k^{'} = 1$) for each original kernel to collaboratively estimate the final response map.
The second structure uses a $k^{'} \times k^{'} \times C \times 1$ filter
(we carefully tune the parameter k' and set k' = k)
shared across all the original filters to calculate the final result.
Both these collaborative layers use less time during inference when compared with the original convolutional layer,
thus they are theoretically able to obtain acceleration.

In many efficient deep learning frameworks such as Caffe~\cite{jia2014caffe},
the convolution operation is reformulated as matrix multiplication by flattening certain dimensions of tensors, such as:
\vspace{-1mm}
{\small
\begin{align}
   V = U^{*} \times W^{*}~~~{\text{s.t.}} ~&~ U^{*} \in R^{XY \times k^{2}C}~,~W^{*} \in R^{k^{2}C \times T}
\end{align}
}
\vspace{-3mm}

\noindent Each row of the matrix $U^{*}$ is related to the spatial position of the output tensor transformed from the tensor $U$,
and $W^{*}$ is a reshaped tensor from weight filters $W$.
These efficient implementations take advantage of the high-efficiency of BLAS libraries,
\eg, GEMM\footnote{matrix-matrix multiplication function} and GEMV\footnote{matrix-vector multiplication function}.

Since each position of the skipped cell in $V^{*}$ corresponds to one row of the matrix $U^{*}$,
we can achieve a realistic speedup in BLAS libraries by reducing the matrix size in the multiplication function.
Different structures of the LCCL need different implementations.
For a $k \times k \times C \times 1$ kernel, the positions of the skipped cells in the original convolutional layer are the same in different channels.
In this situation, we can reduce the size of $U^{*}$ to $S^{'} \times k^{2}C$, where $S^{'}$ is the number of non-zero elements in $V^{'}$.
For a $1 \times 1 \times C \times T$ kernel, the positions of zero cells are different in different channels,
so it is infeasible to directly use the matrix-matrix multiplication function to calculate the result of LCCL, \ie $V^{'}$.
In this case, we have to separate the matrix-matrix multiplication into multiple matrix-vector multiplications.
However, this approach is difficult to achieve the desired acceleration effect.
The unsatisfying acceleration performance of $1 \times 1 \times C \times T$ filters is caused by the inferior efficiency of multiple GEMV,
and some extra operations also cost more time~(\eg, data reconstruction).
Therefore, we choose the $k \times k \times C \times 1$ structure for our LCCL in our experiments,
and leave the acceleration of $1 \times 1 \times C \times T$ filters as our future work.

\subsection{Sparsity Improvement}

According to the previous discussion, the simplest way for model acceleration is directly multiplying the tensor $V^{'}$ and tensor $V$.
However, this approach cannot achieve favourable acceleration performance due to the low sparsity rate of $V^{'}$.

To improve the sparsity of $V^{'}$,
ReLU~\cite{montufar2014number} activation is a simple and effective way by setting the negative values as zeros.
Moreover, due to the redundancy of positive activations,
we can also append $L_1$ loss in the LCCL to further improve the sparsity rate.
In this way, we achieve a smooth $L_{1}L_{2}({\bf X}) = \mu\|{\bf X}\| + \rho|{\bf X}|$ regularizer penalty for each $V^{'}$:
\vspace{-2mm}
{\small
\begin{align}
\|{\bf X}\| = \sqrt{ \sum_{i = 1}^{n} {\bf X}_{i}^2 }~~,~~|{\bf X}| = \sum_{i = 1}^{n} |{\bf X}|
\end{align}
}
\vspace{-2mm}

\noindent However, there are thousands of free parameters in the regularizer term and the additional loss always degrades the classification performance,
as it's difficult to achieve the balance between the classification performance and the acceleration rate.

\begin{table}[ht]
\footnotesize
\begin{center}
\begin{tabular}{ c | c | c | c | c }
\hline
\multirow{2}{*}{Layer} & \multicolumn{2}{| c |}{With BN}   & \multicolumn{2}{| c }{Without BN} \\
                                &      conv1      &      conv2      &      conv1      &      conv2   \\ \hline
res-block-1.2           &      38.8\%     &      28.8\%     &      0.0\%      &      0.0\%   \\ 
res-block-2.2           &      37.9\%     &      23.4\%     &      0.0\%      &      0.2\%   \\
res-block-2.2           &      17.8\%     &      40.4\%     &      0.0\%      &      40.7\%   \\ \hline
\end{tabular}
\end{center}
\caption{Sparsity of the LCCL for different activations with the same training setting.
``With BN'' means we activate the response map of the LCCL by BN and ReLU;
``Without BN'' means we only use ReLU activation.
``x.y'' means the y-th block at x-th stage of ResNet.
We equip six convolutional layers with LCCL on ResNet-20 model.}
\label{table:BN_Sparsity}
\end{table}

Recently, the Batch Normalization (BN)~\cite{ioffe2015batch} is proposed to improve the network performance
and increase the convergence speed during training by stabilizing the distribution and reducing the internal covariate shift of input data.
During this process, we observe that the sparsity rate of each LCCL is also increased.
As shown in Table~\ref{table:BN_Sparsity},
we can find that the BN layer advances the sparsity of LCCL followed by ReLU activation,
and thus can further improve the acceleration rate of our LCCN.
We conjecture that the BN layer balances the distribution of $V^{'}$ and reduces the redundancy of positive values in $V^{'}$ by discarding some redundant activations.
Therefore, to increase the acceleration rate, we carefully integrate the BN layer into our LCCL.

Inspired by the pre-activation residual networks~\cite{he2016identity},
we exploit different strategies for activation and integration of the LCCL.
Generally, the input of this collaborative layer can be either before activation or after activation.
Taking pre-activation residual networks~\cite{he2016identity} as an example,
we illustrate the ``Bef-Aft" connection strategy at the bottom of Fig.~\ref{fig:position_connect}.
``Bef" represents the case that the input tensor is from the flow before BN and ReLU activation.
``Aft" represents the case that the input tensor is the same to the original convolutional layer after BN and ReLU activation.
According to the ``Bef-Aft" strategy in Fig.~\ref{fig:position_connect}.
the ``Bef-Bef", ``Aft-Bef" and ``Aft-Aft" strategies can be easily derived.
During our experiments, we find that input tensors with the ``Bef" strategy are quite diverse when compared with the corresponding convolutional layer due to different activations.
In this strategy, the LCCL cannot accurately predict the zero cells for the original convolutional layer.
So it is better to use the same input tensor as the original convolutional layer, \ie the ``Aft" strategy.

\subsection{Computation Complexity}

Now we analyze the test-phase numerical calculation with our acceleration architecture.
For each convolutional layer, the forward procedure mainly consists of two components,
\ie the low cost collaborative layer and the skip-calculation convolutional layer.
Suppose the sparsity~(ratio of zero elements) of the response map $V^{'}$ is $r$.
We formulate the detailed computation cost of the convolutional layer and compare it with the one equipped with our LCCL.

\begin{table}[ht]
\footnotesize
\begin{center}
\centering
\begin{tabular}{c|c|c}
\hline
Architecture           & FLOPs                   & Speed-Up Ratio   \\\hline
CNN                    & $XYTk^2C$               &  0   \\\hline
basic                  & $XYTC(k{'}^2 + k^2r)$   &  $1 - (k{'}^2/k^2 + r)$\\
($1 \times 1$ kernel)  & $XYTC(1 + k^2r)$        &  $1 - (1/k^2 + r)$ \\
(weight sharing)       & $XYTk^2(1 + Cr)$        &  $1 - (1/C + r)$\\
\hline
\end{tabular}
\end{center}
\caption{Theoretical numerical calculation acceleration for convolutional layers.}
\label{table:theroretical_speedup}
\end{table}

As shown in Table~\ref{table:theroretical_speedup}, the speedup ratio is highly dependent on $r$.
The term $1/C$ costs little time since the channel of the input tensor is always wide in most CNN models and 
it barely affects the acceleration performance.
According to the experiments, the sparsity $r$ reaches a high ratio in certain layers.
These two facts indicate that we can obtain a considerable speedup ratio.
Detailed statistical results are described in the experiments section.

In residual-based networks, if the output of one layer in the residual block is all zero,
we can skip the calculation of descendant convolutional layers and directly predict the results of this block.
This property helps further accelerate the residual networks.

\section{Experiments}

In this section, we conduct experiments on three benchmark datasets to validate the effectiveness of our acceleration method.

\begin{figure*}[!t]
\begin{center}
\includegraphics[width=0.90\textwidth]{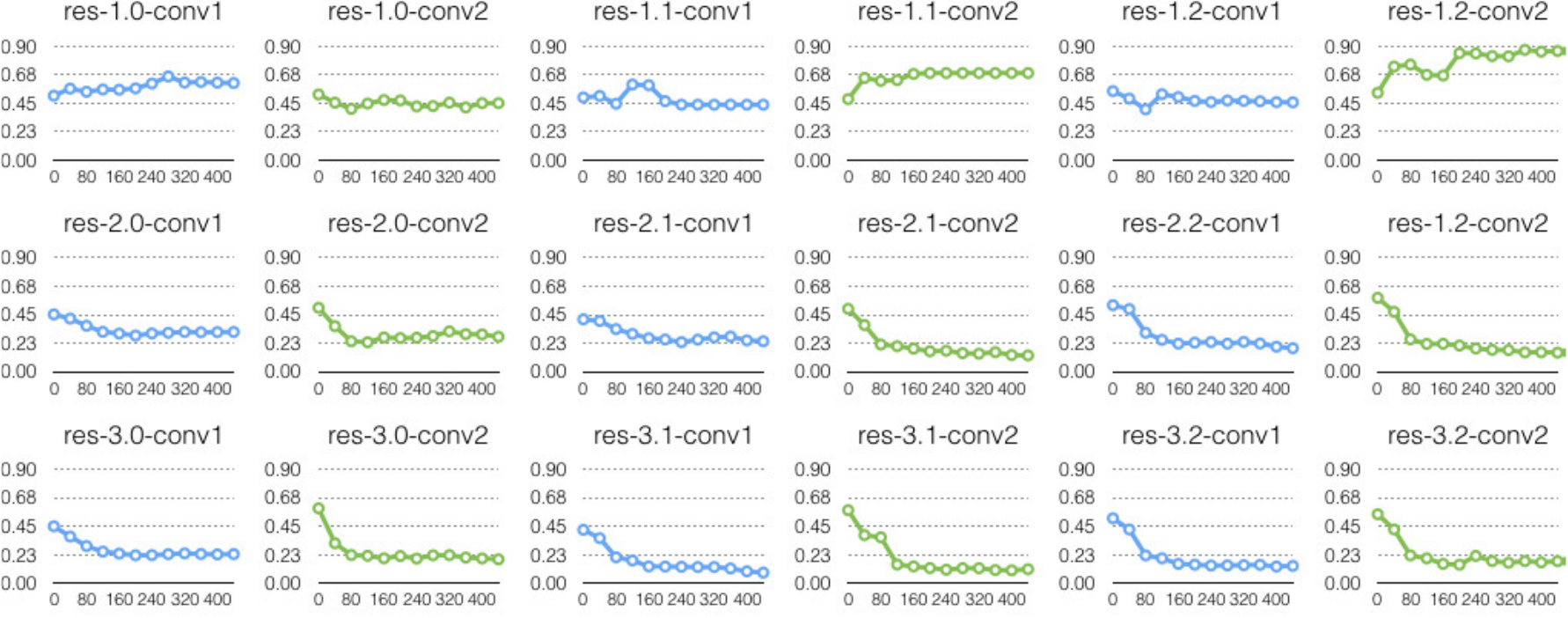}
\end{center}
\caption{Sparsity for the response maps from each collaborative convolutional layer in ResNet-20.
We use LCCL to modify 18 convolutional layers to speed up ResNet-20.
``x.y" represents the y-th residual block in the x-th generalized convolutional block.
``conv1" and ``conv2" represent the first and the second collaboration convolutional in the corresponding residual block.
}
\label{fig:cifar10_sparse}
\end{figure*}

\subsection{Benchmark Datasets and Experimental Setting}

We mainly evaluate our LCCN on three benchmarks: CIFAR-10, CIFAR-100~\cite{krizhevsky2009learning} and ILSVRC-12~\cite{russakovsky2015imagenet}.
The CIFAR-10 dataset contains 60,000 $32 \times 32$ images,
which are categorized into 10 classes and each class contains 6,000 images.
The dataset is split into 50,000 training images and 10,000 testing images.
The CIFAR-100~\cite{krizhevsky2009learning} dataset is similar to CIFAR-10, except that it has 100 classes and 600 images per class.
Each class contains 500 training images and 100 testing images.
For CIFAR-10 and CIFAR-100, we split the 50k training dataset into 45k/5k for validation.
ImageNet 2012 dataset~\cite{russakovsky2015imagenet} is a famous benchmark which contains 1.28 million training images of 1,000 classes.
We evaluate on the 50k validation images using both the top-1 and top-5 error rates.

Deep residual networks~\cite{he2015deep} have shown impressive performance with good convergence behaviors.
Their significance has increased, as shown by the amount of research~\cite{he2016identity,zagoruyko2016wide} being undertaken.
We mainly apply our LCCN to increase the speed of these improved deep residual networks.
In the CIFAR experiments, we use the default parameter setting as~\cite{he2016identity,zagoruyko2016wide}.
However, it is obvious that our LCCN is more complicated than the original CNN model,
which leads to a requirement for more training epochs to converge into a stable situation.
So we increase the training epochs and perform a different learning rate strategies to train our LCCN.
We start the learning rate at 0.01 to warm up the network and then increase it to 0.1 after 3\% of the total iterations.
Then it is divided by 10 at 45\%, 70\% and 90\% iterations where the errors plateau.
We tune the training epoch numbers from \{200, 400, 600, 800, 1000\} according to the validation data

On ILSVRC-12, we follow the same parameter settings as~\cite{he2015deep,he2016identity}
but use different data argumentation strategies.
(1) Scale augmentation: we use the scale and aspect ratio augmentation~\cite{szegedy2015going} instead of the scale augmentation~\cite{simonyan2014very} used in~\cite{he2015deep,he2016identity}.
(2) Color augmentation: we use the photometric distortions from~\cite{howard2013some} to improve the standard color augmentation~\cite{krizhevsky2012imagenet} used in~\cite{he2015deep,he2016identity}.
(3) Weight decay: we apply weight decay to all weights and biases.
These three differences should slightly improve performance (refer to Facebook implementation\footnote{\url{https://github.com/facebook/fb.resnet.torch}}).
According to our experiences with CIFAR, we extend the training epoch to 200, and use a learning rate starting at 0.1 and then is divided by 10 every 66 epochs.

For the CIFAR experiments, we report the acceleration performance and the top-1 error to compare with the results provided in the original paper~\cite{he2016identity,zagoruyko2016wide}.
On ILSVRC-12, since we use different data argumentation strategies, we report the top-1 error of the original CNN models trained in the same way as ours,
and we mainly compare the accuracy drop with other state-of-the-art acceleration algorithms including:
(1) Binary-Weight-Networks~(BWN)~\cite{rastegari2016xnor} that binarizes the convolutional weights;
(2) XNOR-Networks~(XNOR)~\cite{rastegari2016xnor} that binarizes both the convolutional weights and the data tensor;
(3) Pruning Filters for Efficient ConvNets~(PFEC)~\cite{li2016pruning} which prunes the filters with small effect on the output accuracy from CNNs. 

\subsection{Experiments on CIFAR-10 and CIFAR-100}

First, we study the influence on performance of using different connection strategies proposed in the Kernel Selection and Sparsity Improvement sections.
We use the pre-activation ResNet-20 as our base model, and apply the LCCL to all convolutional layers within the residual blocks.
Using the same training strategy, the results of four different connection strategies are shown in Table~\ref{table:res20_connect}.

Both collaborative layers with the after-activation method show the best performance with a considerable speedup ratio.
Because the Aft strategy receives the same distribution of input to that of the corresponding convolution layer.
We also try to use the $L_1L_2$ loss to restrict the output maps of each LCCL.
But this will add thousands of extra values that need to be optimized in the $L_1L_2$ loss function.
In this case, the networks are difficult to converge and the performance is too bad to be compared.

\begin{table}[ht]
\footnotesize
\begin{center}
\begin{tabular}{c|c|c}
\hline
Structure   & Top-1 Err.     & Speed-Up          \\ \hline
Aft-Aft     & \textbf{8.32}  & 34.9\%            \\
Aft-Bef     & 8.71           & 24.1\%            \\
Bef-Bef     & 11.62          & 39.8\%            \\
Bef-Aft     & 12.85          & \textbf{55.4\%}   \\
\hline
\end{tabular}
\end{center}
\caption{Before-activation and after-activation for connection strategy on ResNet-20.
Each LCCL uses $3 \times 3 \times k$ kernel.}
\label{table:res20_connect}
\end{table}

Furthermore, we analyze the performance influenced by using different kernels in the LCCL.
There are two forms of LCCL that collaborate with the corresponding convolutional layer.
One is a tensor of size $1 \times 1 \times C \times T$ (denoted as $1\times1$), and the other is a tensor of size $k \times k \times C \times 1$ (denoted as $k \times k$).
As shown in Table~\ref{table:comparison_kernel},
the $k \times k$ kernel shows significant performance improvement with a similar speedup ratio compared with a $1\times1$ kernel.
It can be caused by that the $k \times k$ kernel has a larger reception field than $1 \times 1$.

\begin{table}[t]
\footnotesize
\begin{center}
\begin{tabular}{ c | c | c | c | c | c | c}
\hline
\multirow{2}{*}{Model} & \multicolumn{3}{| c |}{$1 \times 1 \times C \times T$} & \multicolumn{3}{| c }{$k \times k \times C \times 1$}                  \\
                        & FLOPs & Ratio           & Error       & FLOPs & Ratio           & Error      \\ \hline
ResNet-20               & 3.2E7 & 20.3\%          & 8.57        & 2.6E7 & \textbf{34.9\%} & \textbf{8.32}   \\ 
ResNet-32               & 4.7E7 & \textbf{31.2\%} & 9.26        & 4.9E7 & 28.1\%          & \textbf{7.44}   \\
ResNet-44               & 6.3E7 & \textbf{34.8\%} & 8.57        & 6.5E7 & 32.5\%          & \textbf{7.29}   \\ \hline
\end{tabular}
\end{center}
\caption{Comparison of top-1 error rate on two different collaborative layers.~(The `Ratio' represents the speedup ratio)
}
\label{table:comparison_kernel}
\end{table}

Statistics on the sparsity of each response map generated from the LCCL are illustrated in Fig.~\ref{fig:cifar10_sparse}.
This LCCN is based on ResNet-20 with each residual block equipped with a LCCL configured by a $1 \times 1 \times C \times T$ kernel.
To get stable and robust results, we increase the training epochs as many as possible,
and the sparsity variations for all 400 epochs are provided.
The first few collaborative layers show a great speedup ratio, saving more than 50\% of the computation cost.
Even if the last few collaboration layers behave less than the first few,
the $k \times k \times C \times 1$ based method is capable of achieving more than 30\% increase in speed.

Hitherto, we have demonstrated the feasibility of training CNN models equipped with our LCCL using different low-cost collaborative kernels and strategies.
Considering the performance and realistic implementation, we select the weight sharing kernel for our LCCL.
This will be used in all following experiments as default.

Furthermore, we experiment with more CNN models\cite{he2016identity,zagoruyko2016wide} accelerated by our LCCN on CIFAR-10 and CIFAR-100.
Except for ResNet-164~\cite{he2016identity} which uses a bottleneck residual block
{\tiny
$\left\{
  \begin{array}{ccc}
     1 \times 1 \\
     3 \times 3 \\
     1 \times 1
  \end{array}
\right\}
$
}, all other models use a basic residual block
{\tiny
$\left\{
  \begin{array}{ccc}
     3 \times 3 \\
     3 \times 3
  \end{array}
\right\}
$
}.
We use LCCL to accelerate all convolutional layers except for the first layer, which takes the original image as the input tensor.
The first convolutional layer operates on the original image, and it costs a little time due to the small input channels~(RGB 3 channels).
In a bottleneck structure, it is hard to reach a good convergence with all the convolutional layers accelerated.
The convolutional layer with $1 \times 1$ kernel is mainly used to reduce dimension to remove computational bottlenecks,
which overlaps with the acceleration effect of our LCCL.
This property makes layers with $1 \times 1$ kernel more sensitive to collaboration with our LCCL.
Thus, we apply our LCCL to modify the first and second convolutional layer in the bottleneck residual block on CIFAR-10.
And for CIFAR-100, we only modify the second convolutional layer with $3 \times 3$ kernel in the bottleneck residual block.
The details of theoretical numerical calculation acceleration and accuracy performance are presented in Table~\ref{table:cifar10_acc} and Table~\ref{table:cifar100_acc}.

\begin{table}[t]
\footnotesize
\begin{center}
\begin{tabular}{ c | c | c | c | c}
    \hline
                              & Depth                   & Ori. Err & LCCN    & Speed-up \\\hline
\multirow{2}{*}{ResNet~\cite{he2016identity}}  & 110    & 6.37     & 6.56    & 34.21\%   \\ 
                              & 164*                    & 5.46     & 5.91    & 27.40\%   \\ \hline
\multirow{6}{*}{WRN~\cite{zagoruyko2016wide}}  & 22-8   & 4.38     & 4.90    & 51.32\%  \\
                              & 28-2                    & 5.73     & 5.81    & 21.40\%   \\  
                              & 40-1                    & 6.85     & 7.65    & 39.36\%  \\
                              & 40-2                    & 5.33     & 5.98    & 31.01\%  \\ 
                              & 40-4                    & 4.97     & 5.95    & 54.06\%  \\
                              & 52-1                    & 6.83     & 6.99    & 41.90\%  \\ \hline
\end{tabular}
\end{center}
\caption{Top-1 Error and Speed-Up of eight different CNN models on CIFAR-10~(symbol ``*" means the bottleneck structure).
Ori. Err represents the top-1 error of the original convolution network.}
\label{table:cifar10_acc}
\end{table}

\begin{table}[ht]
\footnotesize
\begin{center}
\begin{tabular}{ c | c | c | c | c}
\hline
                              & Depth                   & Ori. Err     & LCCN    & Speed-up \\\hline
\multirow{1}{*}{ResNet~\cite{he2016identity}}  & 164*   & 24.33        & 24.74   & 21.30\%  \\\hline
\multirow{6}{*}{WRN~\cite{zagoruyko2016wide}}  & 16-4   & 24.53        & 24.83   & 15.19\%  \\
                              & 22-8                    & 21.22        & 21.30   & 14.42\%  \\
                              & 40-1                    & 30.89        & 31.32   & 36.28\%  \\
                              & 40-2                    & 26.04        & 26.91   & 45.61\%  \\ 
                              & 40-4                    & 22.89        & 24.10   & 34.27\%  \\
                              & 52-1                    & 29.88        & 29.55   & 22.96\%  \\ \hline
\end{tabular}
\end{center}
\caption{Top-1 error and speed-up of seven different CNN models on CIFAR-100~(symbol ``*" means the bottleneck structure).
Ori. Err represents the top-1 error of the original convolution network.}
\label{table:cifar100_acc}
\end{table}

Experiments show our LCCL works well on much deeper convolutional networks, such as pre-activation ResNet-164~\cite{he2016identity} or WRN-40-4~\cite{zagoruyko2016wide}.
Convolutional operators dominate the computation cost of the whole network,
which hold more than 90\% of the FLOPs in residual based networks.
Therefore, it is beneficial for our LCCN to accelerate such convolutionally-dominated networks,
rather than the networks with high-cost fully connected layers.
In practice, we are always able to achieve more than a 30\% calculation reduction for deep residual based networks.
With a similar calculation quantity, our LCCL is capable of outperforming original deep residual networks.
For example, on the CIFAR-100 dataset,
LCCN on WRN-52-1 obtains higher accuracy than the original WRN-40-1 with only about 2\% more cost in FLOPs.
Note that our acceleration is data-driven, and can achieve a much higher speedup ratio on ``easy" data.
In cases where high accuracy is not achievable, it predicts many zeros which harms the network structure.

Theoretically, the LCCN will achieve the same accuracy as the original one if we set LCCL as an identity (dense) network.
To improve efficiency, the outputs of LCCL need to be sparse, which may marginally sacrifice accuracy for some cases. 
We also observe accuracy gain for some other cases (WRN-52-1 in Table~\ref{table:cifar100_acc}),
because the sparse structure can reduce the risk of overfitting.

\subsection{Experiments on ILSVRC-12}

We test our LCCN on ResNet-18, 34 with some structural adjustments.
On ResNet-18, we accelerate all convolutional layers in the residual block.
However, ResNet-34 is hard to optimize with all the convolutional layers accelerated.
So, we skip the first residual block at each stage (layer 2, 3, 8, 9, 16, 17, 28, 29) to make it more sensitive to collaboration.
The performance of the original model and our LCCN with the same setting are shown in Table~\ref{table:imagenet_acc}.

\begin{table}[ht]
\footnotesize
\begin{center}
\begin{tabular}{ c | c | c | c | c | c}
\hline
\multirow{2}{*}{Depth} & \multicolumn{2}{| c |}{Top-1 Error} & \multicolumn{2}{| c |}{Top-5 Error} & \multirow{2}{*}{Speed-up}\\
                       &      ResNet      &     LCCN   &      ResNet      &     LCCN   &                          \\\hline
18                     &      30.02       &     33.67  &      10.76       &     13.06  &  34.6\%                  \\\hline
34                     &      26.58       &     27.01  &      8.64        &     8.81   &  24.8\%                  \\\hline
\end{tabular}
\end{center}
\caption{Top-1 and Top-5 Error of LCCN on ImageNet classification task.}
\label{table:imagenet_acc}
\end{table}

We demonstrate the success of LCCN on ResNet-18,~34~\cite{he2016identity},
and all of them obtain a meaningful speedup with a slight performance drop.

\begin{table}[ht]
\footnotesize
\begin{center}
\begin{tabular}{ c | c | c | c | c}
    \hline
    Depth            & Approach & Speed-Up            & Top-1 Acc. Drop  & Top-5 Acc. Drop \\\hline
\multirow{3}{*}{18}  & LCCL     & 34.6\%              & 3.65             &   2.30      \\
                     & BWN      & $\approx 50.0\%$    & 8.50             &   6.20      \\
                     & XNOR     & $\approx 98.3\%$    & 18.10            &   16.00     \\\hline
\multirow{2}{*}{34}  & LCCL     & 24.8\%              & 0.43             &   0.17     \\
                     & PFEC     & 24.2\%              & 1.06             &   -        \\\hline
\end{tabular}
\end{center}
\caption{Comparison with other acceleration methods on ResNet.
Acc. Drop represents the accuracy drop.}
\label{table:compare_acc_18}
\end{table}

We compare our method with other state-of-the-art methods, shown in Table~\ref{table:compare_acc_18}.
As we can see, similar to other acceleration methods, there is some performance drop.
However, our method achieves better accuracy than other acceleration methods.

\subsection{Theoretical vs. Realistic Speedup}

There is often a wide gap between theoretical and realistic speedup ratio.
It is caused by the limitation of efficiency of BLAS libraries, IO delay, buffer switch or some others.
So we compare the theoretical and realistic speedup with our LCCN.
We test the realistic speed based on Caffe~\cite{jia2014caffe}, an open source deep learning framework.
OpenBLAS is used as the BLAS library in Caffe for our experiments.
We set CPU only mode and use a single thread to make a fair comparison.
The results are shown in Table~\ref{table:Comparison_Speed}.

\begin{table}[ht]
\footnotesize
\begin{center}
\begin{tabular}{ c | c | c | c  | c | c | c }
    \hline
 \multirow{2}{*}{Model} & \multicolumn{2}{| c |}{FLOPs} & \multicolumn{2}{| c |}{Time (ms)} & \multicolumn{2}{| c }{Speed-up} \\
             & CNN   & LCCL  & CNN   & LCCL  & Theo    & Real    \\ \hline
ResNet-18    & 1.8E9 & 1.2E9 & 97.1  & 77.1  & 34.6\%  & 20.5\%  \\ \hline
ResNet-34    & 3.6E9 & 2.7E9 & 169.3 & 138.6 & 24.8\%  & 18.1\%  \\ \hline
\end{tabular}
\end{center}
\caption{Comparison on the theoretical and realistic speedup.}
\label{table:Comparison_Speed}
\end{table}


\textbf{Discussion.}
As shown in Table~\ref{table:Comparison_Speed}, our realistic speedup ratio is less than the theoretical one,
which is caused mainly by two reasons.
First, we use data reconstruction and matrix-matrix multiplication to achieve the convolution operator as Caffe~\cite{jia2014caffe}.
The data reconstruction operation costs too much time, making the cost of our LCCL much higher than its theoretical speed.
Second, the frontal convolution layers usually take more time but contain less sparsity than the rear ones,
which reduces the overall acceleration effect of the whole convolution neural network.
These two defects can be solved in theory, and we will focus on the realistic speedup in future.

\textbf{Platform.}
The idea of reducing matrix size in convolutional networks can be applied to GPUs as well in principle, 
even though some modifications on our LCCN should be made to better leverage the existing GPU libraries.
Further, our method is independent from platform, and should work on the FPGA platform with customization.

\subsection{Visualization of LCCL}

\begin{figure}[ht]
\begin{center}
\includegraphics[width=0.45\textwidth]{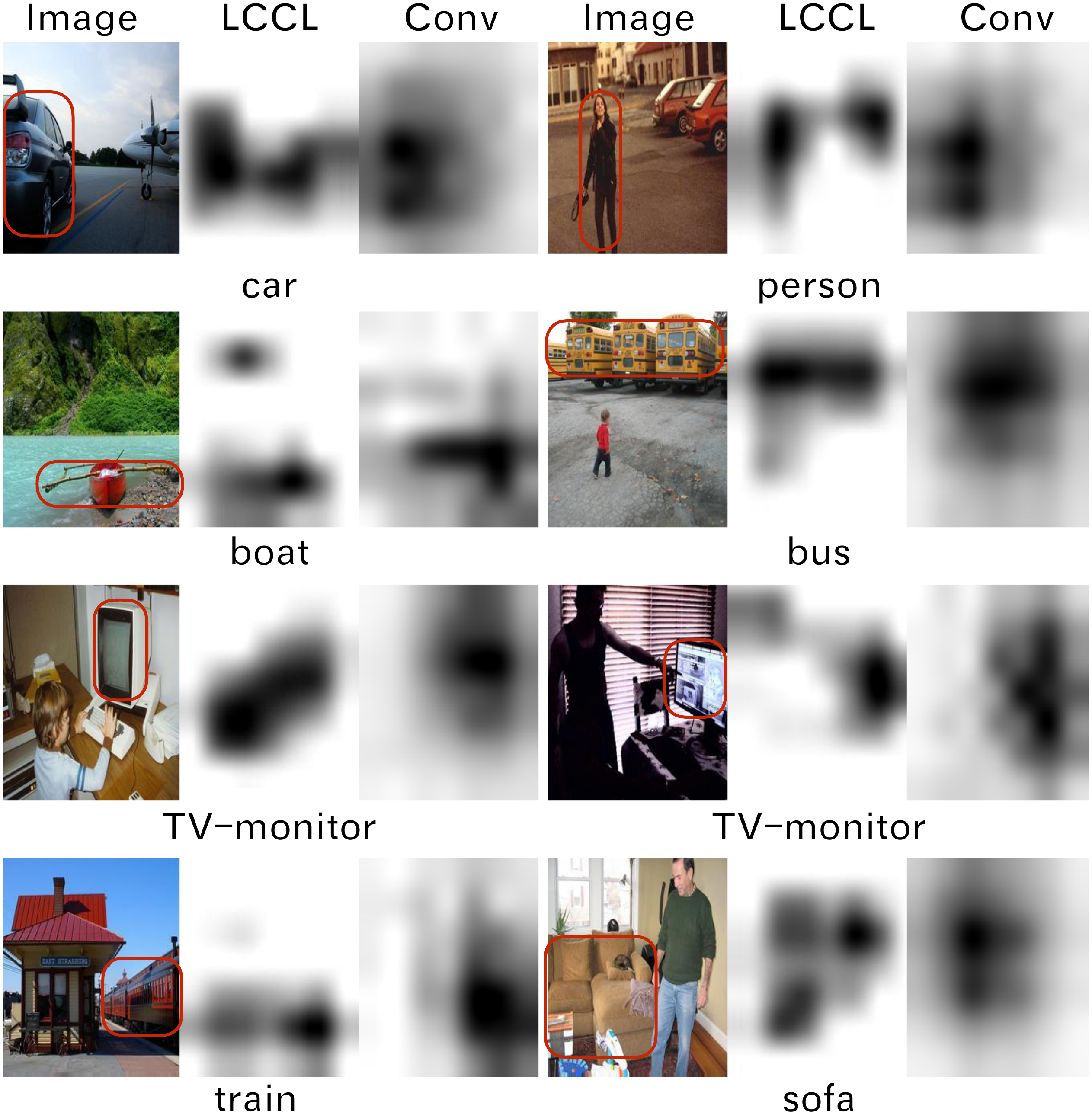}
\end{center}
\caption{
The feature maps (after ReLU) generated from the last LCCL of our LCCN and the corresponding convolutional layer of ResNet-50 are visualized for testing samples of PASCAL VOC2007 dataset.
Each triplet represents one picture and its corresponding feature maps.
The activated area of LCCL seems highlight more foreground objects than that of ResNet-50.
In the meantime, LCCL is possible to depress the background area.
}
\label{fig:response_map}
\end{figure}

Here is an interesting observation about our LCCL.
We visualize the results of LCCN on PASCAL VOC2007~\cite{pascal-voc-2007} training dataset.
We choose ResNet-50 as the competitor, and add an additional 20 channels' convolutional layer with an average pooling layer as the classifier.
For our LCCN, we equip the last 6 layers of this competitor model with our LCCL.
After fine tuning,
the feature maps generated from the last LCCL and the corresponding convolutional layer of the competitor model are visualized in Fig.~\ref{fig:response_map}.
As we can observe, our LCCL might have the ability to highlight the fields of foreground objects, and eliminates the impact of the background via the collaboration property.
For example, in the second triplet, car and person are activated simultaneously in the same response map by the LCCL.

At the first glance, these highlighted areas look similar with the locations obtained by attention model.
But they are intrinsically different in many ways,
\eg, motivations, computation operations, response meaning and structures.

\section{Conclusion}

In this paper, we propose a more complicated network structure yet with less inference complexity to accelerate the deep convolutional neural networks.
We equip a low-cost collaborative layer to the original convolution layer.
This collaboration structure speeds up the test-phase computation by skipping the calculation of zero cells predicted by the LCCL.
In order to solve the the difficulty of achieving acceleration on basic LCCN structures,
we introduce ReLU and BN to enhance sparsity and maintain performance.
The acceleration of our LCCN is data-dependent,
which is more reasonable than hard acceleration structures.
In the experiments, we accelerate various models on CIFAR and ILSVRC-12, 
and our approach achieves significant speed-up, with only slight loss in the classification accuracy.
Furthermore, our LCCN can be applied on most tasks based on convolutional networks
(\eg, detection, segmentation and identification).
Meanwhile, our LCCN is capable of plugging in some other acceleration algorithms
(\eg, fix-point or pruning-based methods), which will further enhance the acceleration performance.

{\small
\bibliographystyle{ieee}
\bibliography{More_is_Less}
}

\end{document}